\documentclass{ifacconf}

\usepackage{graphicx}      
\usepackage[utf8]{inputenc}
\usepackage{multirow}
\usepackage{amsmath,bm}
\usepackage{array}
\usepackage{upgreek}
\usepackage{natbib}

\makeatletter
\def\endfigure{\end@float}
\def\endtable{\end@float}
\makeatother

\usepackage[caption=false]{subfig}
\usepackage{etoolbox}

\AtBeginEnvironment{figure}{\setlength{\captionwidth}{0.8\linewidth}}
\AtBeginEnvironment{table}{\setlength{\captionwidth}{0.8\linewidth}}
\begin{document}
\begin{frontmatter}

\title{Modeling and Soft-fault Diagnosis of Underwater Thrusters with Recurrent Neural Networks} 

\thanks[footnoteinfo]{This work was supported by FP7-PEOPLE-2013-ITN project (ROBOCADEMY), funded by the European Commission.}

\author[First]{Samy Nascimento} 
\author[First]{Matias Valdenegro-Toro} 

\address[First]{German Research Center for Artificial Intelligence, 
   Bremen, 28359 DE (e-mail: samy.marcelo\_nascimento@dfki.de, matias.valdenegro@dfki.de).}
   

\begin{abstract}                
Noncritical soft-faults and model deviations are a challenge for Fault Detection and Diagnosis (FDD) of resident Autonomous Underwater Vehicles (AUVs). 
Such systems may have a faster performance degradation due to the permanent exposure to the marine environment, and constant monitoring of component conditions is required to ensure their reliability.
This works presents an evaluation of Recurrent Neural Networks (RNNs) for a data-driven fault detection and diagnosis scheme for underwater thrusters with empirical data.
The nominal behavior of the thruster was modeled using the measured control input, voltage, rotational speed and current signals. We evaluated the performance of fault classification using all the measured signals compared to using the computed residuals from the nominal model as features.

\end{abstract}


\end{frontmatter}

\section{Introduction}
\label{sec:intro}
There are increasing needs to monitor underwater facilities and structures, such as communication lines, wind farms and oil rigs. Autonomous Underwater Vehicles (AUVs) can perform this work with less risk to human lives, leading to an increased interest on sub-sea resident technology. These systems are meant to operate in a harsh environment with little to none access by operators, thus being required to perform several tasks of self-inspection autonomously. This work is part of the development and evaluation of methods to enable autonomous Fault Detection and Diagnosis (FDD) to enhance the self-awareness of sub-sea resident AUVs.

Faults can be divided into two major categories: Hard-faults and Soft-faults. Hard-faults happen abruptly, such as a propeller blade breaking, or rotor blocking. In contrast, Soft-faults are characterized by a continuous deviation of a property from normal conditions, and usually builds up over time, leading to a performance degradation of the affected component.
Such deviations are hard to detect at early stages, and may be not critical for the normal operation of the system, but preemptive monitoring of its development is needed. Propeller biofouling and corrosion due to sea water are examples of Soft-faults inducing processes.

Several methods have been proposed for the soft-fault detection and compensation, and most of the approaches depend on explicit physical models of the plant and the faults. Observers, Extended Kalman Filters (EKF) and parameter estimation are examples of such methods.
One of the main advantages of data-driven approaches is the ability of learning specific features of the signals generated by the plant, requiring less knowledge of the exact nature or extension of the fault. In the specific case of underwater thrusters, their non-linear dynamic nature might require complex models that are hard to identify with classical methods. Estimation of component condition to perform diagnosis of thruster soft-faults and noncritical failures is challenging, due to the amount and nature of external and internal factors that affect its performance. However, an accurate diagnosis is required to enable better response to the occurrence of these faults on a real scenario.

Neural Networks are a well-proven method for nonlinear dynamic system identification, being a computationally efficient method for nonlinear regression. Several configurations were evaluated for this task \citep{Narendra1990, Chen1990} and have been used in many industrial applications, including control and fault diagnosis.  \cite{Samy2011} presents a Neural Network based fault detection and isolation diagnosis system for an aerial vehicle.
More recently, the works of \cite{Ogunmolu2016} and \cite{Wang2017} present new contributions for dynamic system modeling using RNNs and Deep Learning, with the use of Long Short-term Memory (LSTM) networks.

In a previous work \citep{Nascimento2017}, we evaluated several regressors for online modeling of thrusters under changing conditions. This approach uses an initial model as reference to the nominal operation of the component and an adaptive model able to learn non-critical performance deviations. The rotational speed and current signals are used for fault detection, in a similar approach as presented on the work of \cite{Caccia2001}.  In this paper, we extend this evaluation to the diagnosis of non-critical soft-faults using a fault classifier.

\section{RNN-based Modeling}
\label{sec:models}

In this section, we shortly review some of Recurrent Neural Networks (RNNs) topologies that are used in this work. The Nonlinear Autoregressive Exogenous Model (NARX) networks are a classic scheme to identify dynamic systems. The Long Short-term Memory (LSTM) cell was introduced to solve the problem of gradient vanishing in normal RNNs and the Gated Recurrent Unit (GRU) was introduced recently as a smaller and simpler recurrent cell. 

\subsection{Nonlinear Autoregressive Exogenous Model}

A Nonlinear Autoregressive Exogenous Model derives from the expansion of the past values of the inputs and outputs (targets) of an estimator to be used use as extra features. This expansion creates a time dependency and enables the learning of certain dynamic behavior.

A nonlinear autoregressive exogenous model is derived from the use of past values of input and output. Given the input vector $u(t) = \{u_1(t), u_2(t), \ldots, u_n(t)\}$ and
the output vector $y(t) = \{y_1(t), y_2(t), \ldots, y_m(t)\}$, a NARX model output is defined by:

\begin{multline}
y(t) = f(y(t-1), y(t-2), \ldots, y(t-n), \ldots, \\
u(t-1), u(t-2), \ldots, u(t-q)) + \varepsilon_{t}
\end{multline}

Applying such a scheme to NNs, it is possible to performing estimation of nonlinear relations based on current and previous values. NARX networks can learn dynamic behavior and preserve long-term dependencies longer than a standard RNN.

\begin{figure}[!ht]
\subfloat[\label{fig:narx-sp}]{%
\includegraphics[scale=0.5]{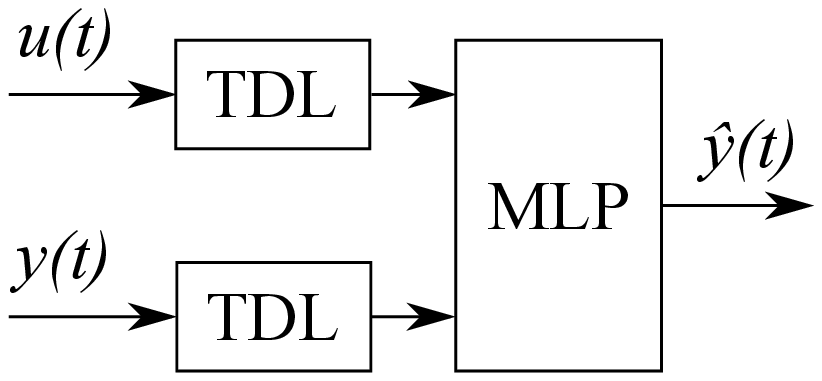}
}
\hfill
\subfloat[\label{fig:narx-p}]{%
\includegraphics[scale=0.5]{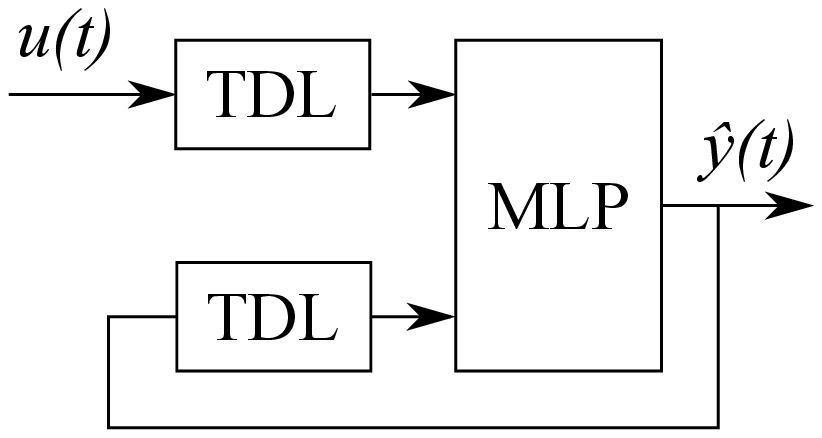}
}
\caption{\label{fig:narx}Nonlinear Auto Regressive Exogenous Model Networks. Series-parallel (a) and parallel (b) configurations are shown. TDL denotes a tapped delay line applied to the input or the output vector.}
\end{figure}

\subsection{Deep Learning Recurrent Networks}

Long Short-term Memory networks is a type of recurrent neural networks proposed in \cite{Hochreiter1997}.
With the introduction of a memory cell and the input, output and forget gates, such networks do not present the vanishing gradient problem and are able to preserve information for longer periods.

\begin{figure}[!htb]
\centering
\includegraphics[width=0.4\linewidth]{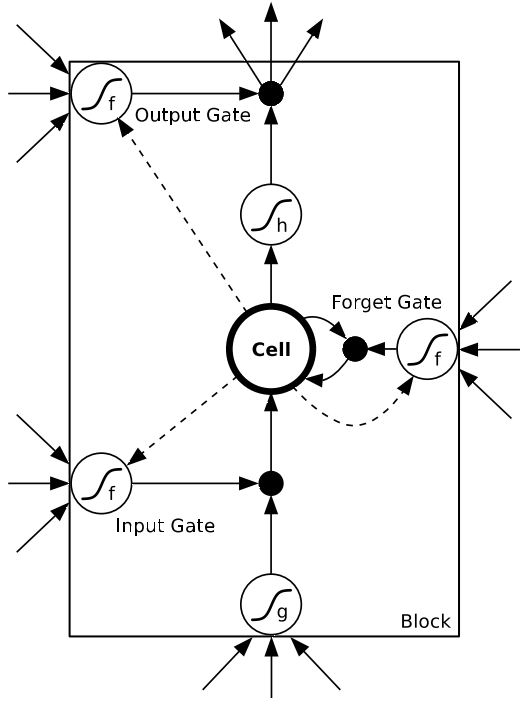}
\caption{\label{fig:lstm}Expanded LSTM network cell operation \cite{Graves2012}.}
\end{figure}

\begin{align}
f_t &= \sigma_g(W_{f} x_t + U_{f} h_{t-1} + b_f) \\
i_t &= \sigma_g(W_{i} x_t + U_{i} h_{t-1} + b_i) \\
o_t &= \sigma_g(W_{o} x_t + U_{o} h_{t-1} + b_o) \\
c_t &= f_t \circ c_{t-1} + i_t \circ \sigma_c(W_{c} x_t + U_{c} h_{t-1} + b_c) \\
h_t &= o_t \circ \sigma_h(c_t)
\end{align}

In the above described cell, $x_t$ is the input vector, $f_t$ is the forget gate activation vector, 
$i_t$ represents the input gate activation vector, $o_t$ is the output gate activation vector and $h_t$ is output vector. $W$, $U$ and $b$ represent respectively weight matrices and bias vector.

Presented by \cite{Cho2014}, Gated Recurrent Units are a special case of the LSTM cell, introduced to be computationally cheaper and easier to train. The reader is encouraged to refer to \cite{Goodfellow2017} to a complete description of Deep Learning techniques.

\section{Data-driven Thruster Modeling}
\label{sec:thruster_modeling}

\subsection{Data collection}
\label{sec:data_collection}

\begin{figure}[!ht]
\centering
\subfloat[\label{fig:t100_thruster}]{%
\includegraphics[height=0.30\linewidth]{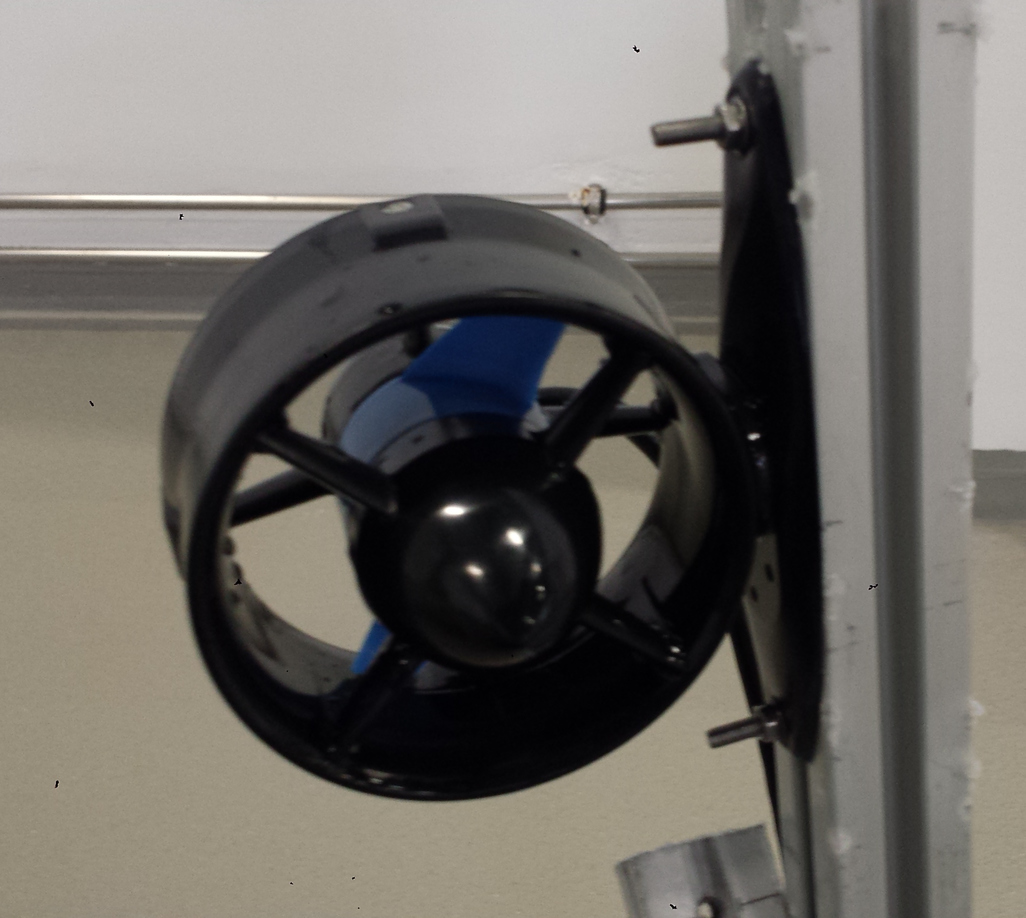}
}
\hfill
\subfloat[\label{fig:graupner_controller}]{%
\includegraphics[height=0.30\linewidth]{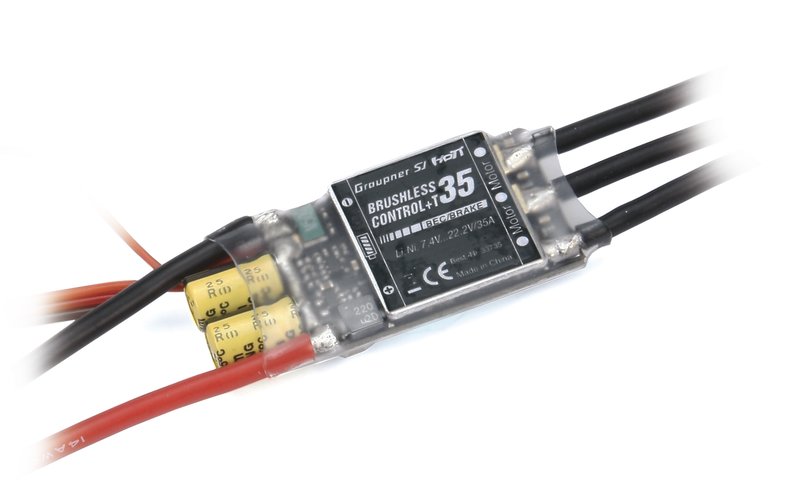}
}
\caption{T100 thruster. T35 controller (Source: Graupner).}
\label{fig:controller-thruster}
\end{figure}

Data was collected with several Bluerobotics T100 thrusters.
This model is composed by an electric brushless motor, ranging from 300 to 4200 rpm,
has up to 130 W of output power and has 2.36 kgf of nominal torque (Figure \ref{fig:t100_thruster}).
The thrusters were driven by a Graupner T35 electronic speed controller (ESC), which provides the signals of rotational speed, current, voltage and temperature and is rated to 35 A (Figure \ref{fig:graupner_controller}). An ATMEGA328 microcontroller was used as interface between the supervisory software running on a computer and the ESC. 

A sequence of experiments were performed to model the thruster behavior in several conditions and evaluate a classification-based diagnosis approach. The dataset was obtained with a testbed using Bluerobotics T100 thrusters and controllers capable of giving rotational speed, voltage and current consumption. Although the relation of thrust and rotational speed is given by the manufacturer, we did not consider this indirect thrust measurement as an output variable of the model.

\begin{figure}[t]
\centering
\includegraphics[width=0.60\linewidth]{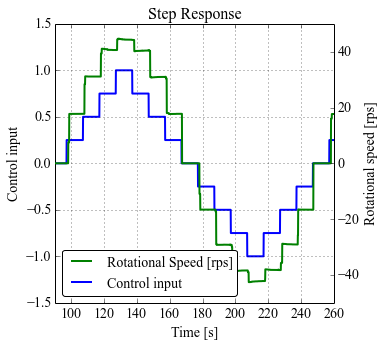}
\caption{Step response of the thruster for multiple steps of 0.25 of the maximum input. The system presents a second-order-like response with average deadtime of 0.59 seconds and average settling time of 2.95 seconds.\label{fig:step-overtime}}
\end{figure}

\begin{figure}[ht]
\centering
\includegraphics[width=0.60\linewidth]{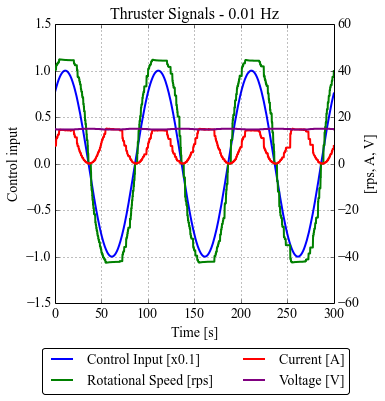}
\caption{Thruster response over time to a sinusoidal signal of 0.01 Hz.
\label{fig:signals-u1-overtime}}
\end{figure}

\begin{figure}[ht]
\centering
\includegraphics[width=0.60\linewidth]{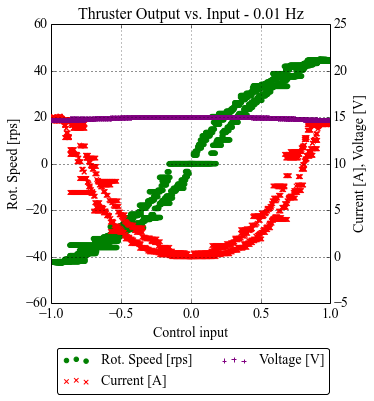}
\caption{Scatter plot of the thruster response to a sinusoidal signal of 0.01 Hz.\label{fig:signals-u1-scatter}}
\end{figure}

\begin{figure}[ht]
\includegraphics[width=\linewidth]{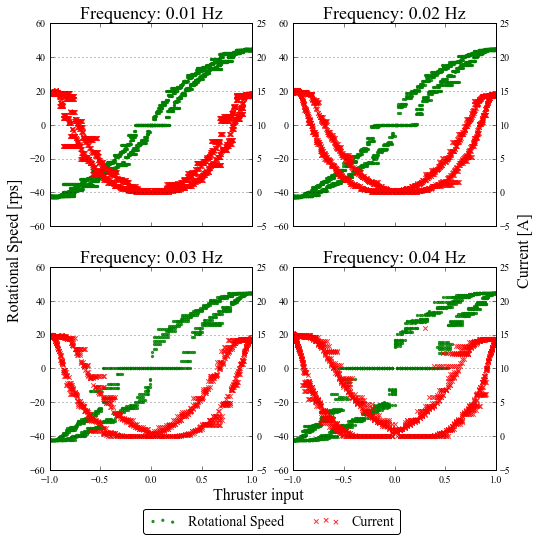}
\caption{Response to input signals of different frequencies. These curves show a hysteresis caused by the intrinsic delay of the controller-thruster system.\label{fig:frequency-curves}}
\end{figure}

The responses of the controller-thruster assembly to a step and sinusoidal signal in open loop were used to collect data for the model training. These signals represent a pulse width modulation signal with width varying from 1.0 to 2.0 milliseconds, mapped to the -1.0 to 1.0 range to represent direction. For the step response characterization, every step of the applied input had 0.25 of the maximum possible amplitude, as shown in Figure \ref{fig:step-overtime}. For the general model, data was obtained using sinusoidal input signals with of 0.01, 0.02, 0.03 and 0.04 Hz. In the Figures \ref{fig:signals-u1-overtime}, \ref{fig:signals-u1-scatter} and \ref{fig:frequency-curves}, the characteristic thruster open loop response is shown, with a saturated curve for the rotational speed and a quadratic curve for the current. 

The step response in open-loop for identification, depicted in Fig. \ref{fig:step-overtime}, showed a second-order overdamped system with some dead-time. The averaged measurements for settling time were 2.95 seconds and 0.59 seconds for dead-time. Also, according to the manufacturer, the thruster presents a dead band of $\pm$ 25 $\upmu$s, which was considered for further modeling. Due to the effect of the dead time and deadband, the system presents an observable hysteresis, that may impact on the selection of the frequency of signal to perform modeling of the thruster, as shown in Fig. \ref{fig:frequency-curves}.

\subsection{Nominal Model Estimation}
\label{subsec:nominal_model}

Four NN-based methods were evaluated for the role of identifying the nominal model: Multilayer Perceptron (MLP) as a baseline regressor, a NARX Neural Networks, LSTM-based network and GRU-based network. The inputs to each model are control signal and voltage, while the outputs are current and rotational speed.

Since purely static regressors are not able to model the characteristic dead-time of the thruster, for the MLP, the known dead-time delay and the dead band non-linearities were added to the estimator in a Hammerstein-Wiener model fashion. For the NARX network, only the deadband was added and their performances evaluated.
 
Scaling the data vectors was required for training the networks, since this strongly affects the performance of the regressors. Inputs and targets were scaled by the Interquartile Range (IQR), wich allows some outlier robustness.
Grid search was used to tune the parameters of each regressor, as shown on Table \ref{tab:reg_config}. Besides the number of neurons on every configuration, the batch size and the numbers of past time steps to be taken into training (Lookback) were adjusted. For the NARX networks, every input and target feature was considered to have independent tapped delay line operators.
The training set consisted of 2000 samples of the sinusoidal signal with 0.01 Hz. The regressors were trained with a 2-fold time series cross validation scheme, in which the validation set is always taken from a section posterior than the training set. The regressors were compared with a 1000 sample dataset in each frequency analyzed (0.01, 0.02, 0.03 and 0.04 Hz). 

\begin{table}[ht]
\begin{center}
\footnotesize
\caption{Configurations and scores for the best evaluated regressors after grid search. Training set: 2000 samples. Test set: 1000 samples. Notation: L for LSTM cell, G for gated recurrent unit, P for perceptron, TDL for time delay, DB: deadband. Activation: tanh, optimizer: ADAM.\label{tab:reg_config}}
\footnotesize
\begin{tabular}{c >{\centering}m{1.5cm} >{\centering}m{1.5cm} c}
\multicolumn{4}{c}{\textbf{Regressors - Configurations}}\\
\hline
\textbf{Method} & \textbf{Hidden Layers} & \textbf{Batch Size} & \textbf{\centering Lookback} \\ 
\hline
MLP                 & 8P,4P         & 10 & -            \\
MLP + TDL + DB      & 8P,4P         & 10 & -            \\ 
NARX NN             & 32P,4P        & 5  & (20,0),(2,0) \\
NARX NN + DB  		& 32P,4P   		& 5  & (20,0),(2,0)      \\ 
LSTM		        & 24L,16P,16P & 5  & 1            \\
GRU                 & 24G,16P,16P & 5  & 1            \\
\hline 
\end{tabular}
\end{center}
\end{table}

\begin{figure}[ht]
\includegraphics[width=\linewidth]{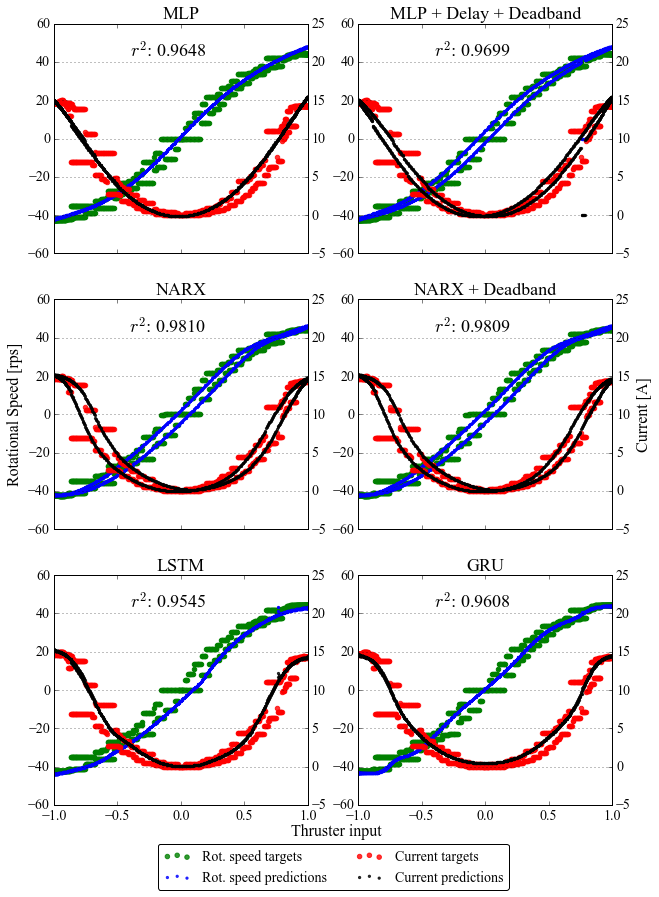}
\caption{Regression results for several regressors for a 2000/1000 train/test dataset. The NARX networks present a $r^2$-score of 0.9809, thus having the best performance in this case.\label{fig:regression-results}}
\end{figure}

\begin{figure}[ht]
\includegraphics[width=\linewidth]{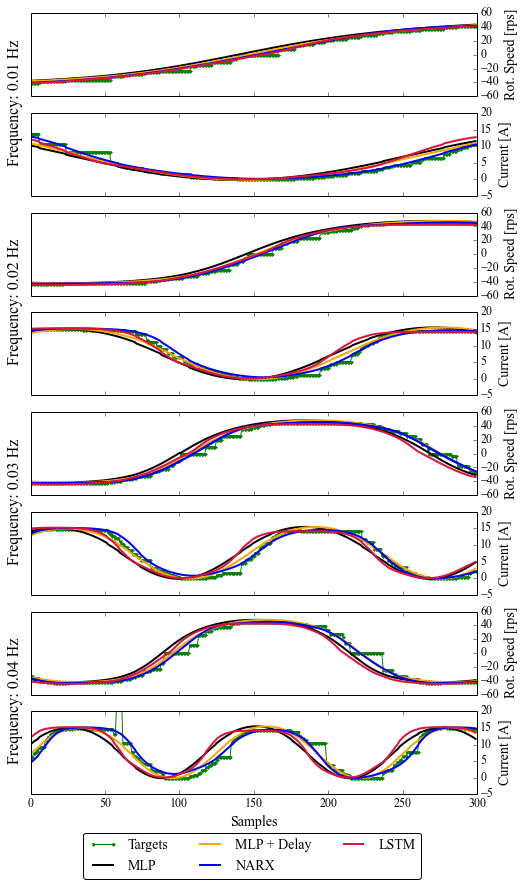}
\caption{Regressor curves for estimation of thruster rotational speed and current for a input sinusoidal signals of several frequencies.\label{fig:frequency-overtime}}
\end{figure}

The mean coefficients of determination ($r^2$ scores) between current and rotational speed outputs, for every frequency, are shown in Table \ref{tab:r2-scores}. The scatter plot for a 1000 sample test set of a sinusoid of 0.01 Hz is shown in the Figure \ref{fig:regression-results}. The results over time for 300 sample test set for the different frequencies analysed are shown in the Figure \ref{fig:frequency-overtime}. 

\begin{table}[ht]
\centering
\caption{\label{tab:r2-scores}Coefficient of determination ($r^2$-score) for the the evaluated regressor for test sets with several frequencies. The shown scores are the average of 30 runs. The results for the 0.04 Hz signal are contaminated by outliers, thus presenting lower regression scores. D: time delay, DB: deadband.}
\footnotesize
\begin{tabular}{ccccc}
\multicolumn{5}{c}{\textbf{Regression Scores - Frequency}}\\
\hline
\multirow{2}{1cm}{\textbf{Method}} & \multicolumn{4}{c}{\textbf{$r^2$ Scores}}\\
\cline{2-5}
& \textbf{0.01 Hz} & \textbf{0.02 Hz} & \textbf{0.03 Hz} & \textbf{0.04 Hz}\\ 
\hline
MLP       & 0.9650 & 0.9580  & 0.9139 & 0.5340 \\
MLP+D+DB  & 0.9794 & 0.9867  & 0.9694 & 0.6260 \\ 
NARXNN	  & 0.9745 & 0.9931  & 0.9579 & 0.6432 \\
NARXNN+DB & 0.9746 & 0.9931  & 0.9579 & 0.6431 \\ 
LSTM		  & 0.9563 & 0.9542  & 0.9058 & 0.5224 \\
GRU       & 0.9590 & 0.9585  & 0.9112 & 0.5258 \\
\hline 
\end{tabular}
\end{table}

The results of Table \ref{tab:r2-scores} and Figures \ref{fig:regression-results} and \ref{fig:frequency-overtime} show that the hysteresis caused by time delay were better captured by the NARX network. This network sctructure is also able to rubustly predict the system response to other input signal frequencies. This regressor was then used as a reference nominal model for the fault classification task (see Section \ref{sec:classification}).

\section{Soft-Fault Classification}
\label{sec:classification}

\begin{figure}[ht]
\subfloat[\label{subfig:t100_broken}]{%
\includegraphics[height=0.4\linewidth]{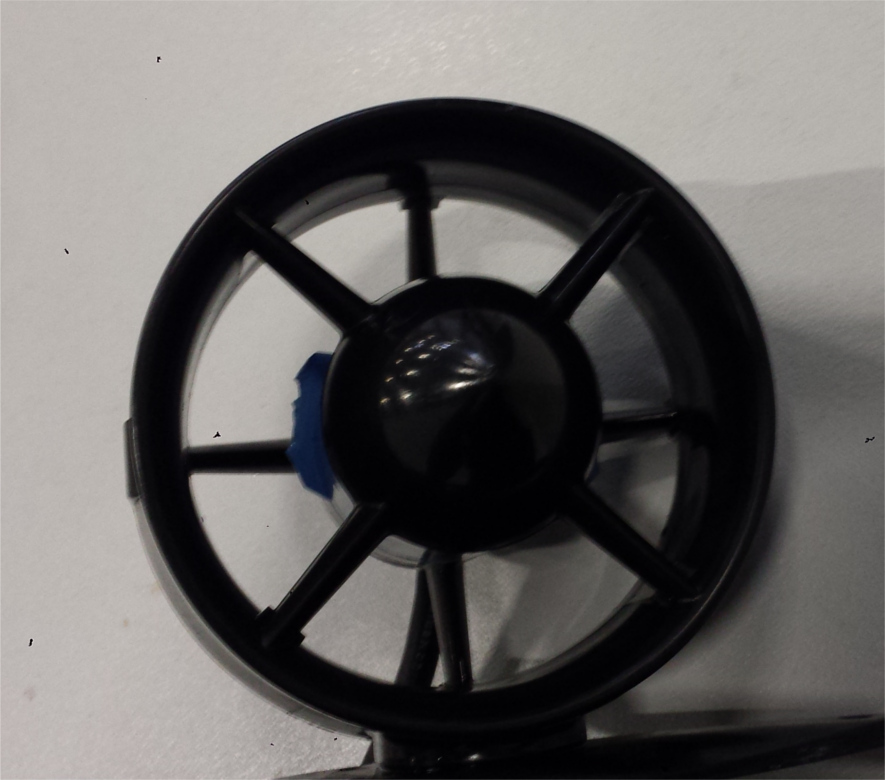}
}
\hfill
\subfloat[\label{subfig:t100_silicon}]{%
\includegraphics[height=0.4\linewidth]{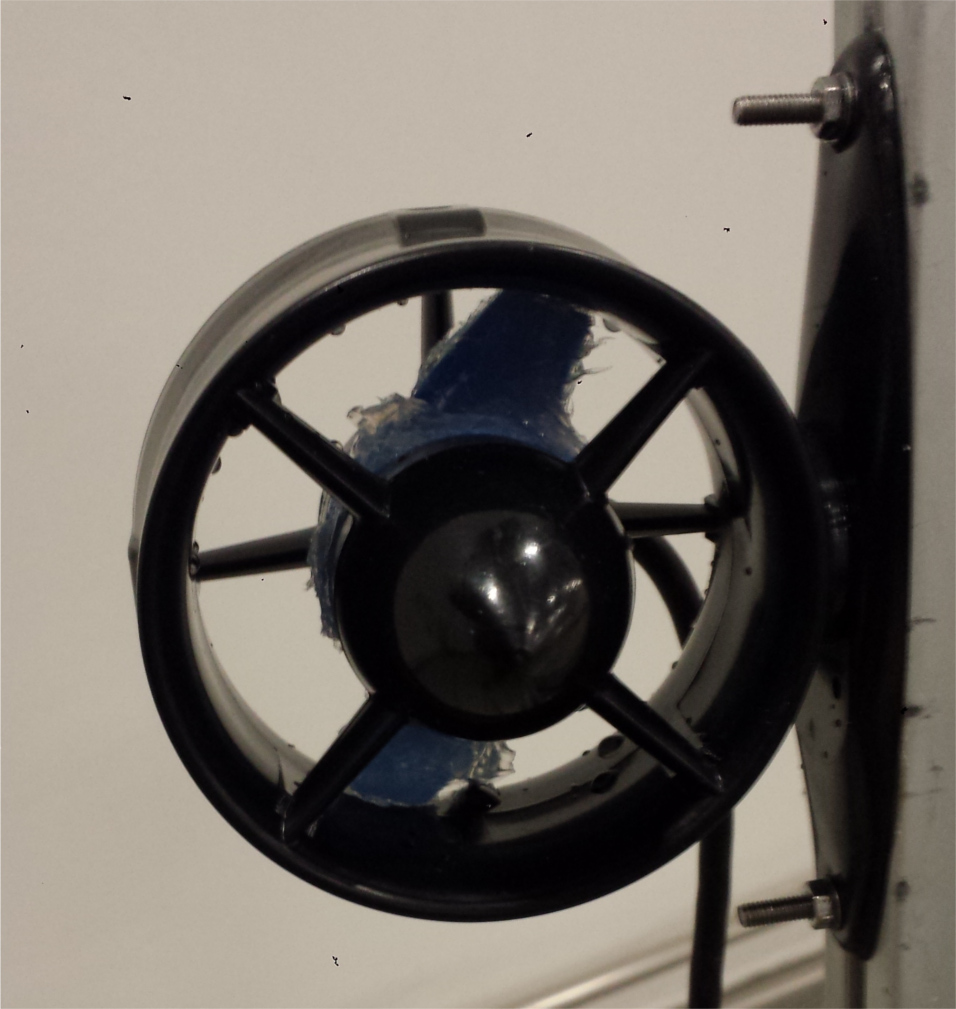}
}
\caption{Thruster with (a) broken propellers and (b) impregnated with silicon.\label{fig:conditions}}
\end{figure}

\begin{figure}[ht]
\includegraphics[width=\linewidth]{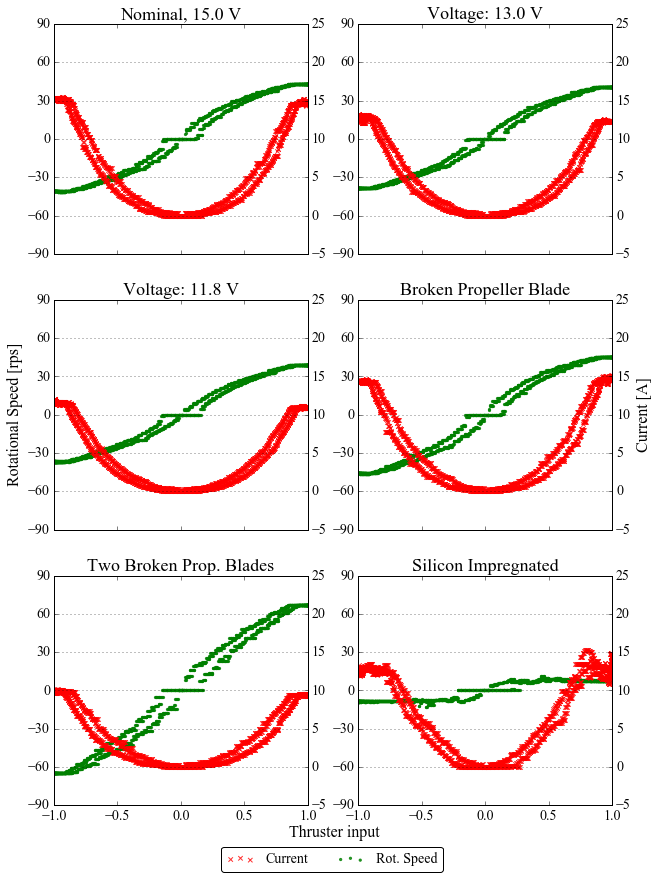}
\caption{Datasets for different conditions, with 2000 samples each. Data is shown for normal operation with supply voltages of 15.0, 13.0 and 11.8 V.\label{fig:conditions-datasets}}
\end{figure}

\begin{figure}[ht]
\includegraphics[width=\linewidth]{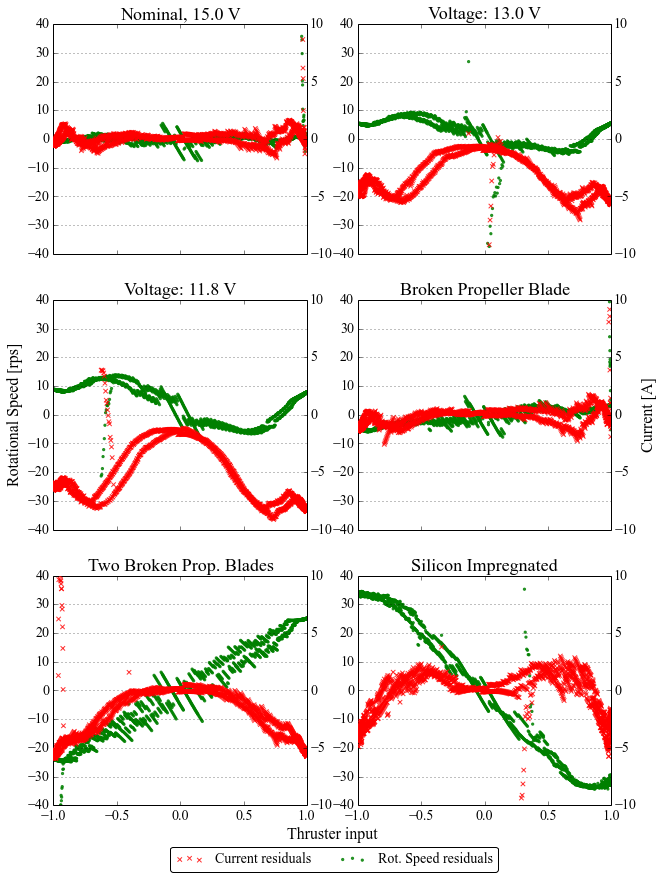}
\caption{Computed residuals for each thruster condition.\label{fig:conditions-residuals}}
\end{figure}

For the task of fault diagnosis, data was collected and labeled with a signal of 0.01 Hz for the following operational conditions: nominal operation with 15.0 V, 13.0 V, 11.8 V, with one and two broken propeller blades and with a propeller impregnated with silicon to simulate biofouling (see Figure \ref{fig:conditions}). A distinction between model changes due to healthy low-voltage states and faulty states may lead to a better fault handling decision making. These conditions are represented by six classes in our fault classifier. The scatter plots for each condition are depicted in Figure \ref{fig:conditions-datasets}. The residuals calculated from comparison to the nominal model obtained previously (NARX) were used for the fault classification task as features and are shown in the Figure \ref{fig:conditions-residuals}.

Several classifiers were trained with each condition as a class, in two configurations. The first configuration uses the thruster input, voltage, rotational speed and current signals as features, with 2000 samples for each class. The second configuration uses the rotational speed and current residuals as features using 2000 residual samples each. Two methods were evaluated for diagnosis: multilayer perceptron and LSTM-based classifers in the configurations listed on the Table \ref{tab:classifiers-results}. A 3-fold time series split cross-validation scheme and cross entropy loss function were used for training of the classifiers. A test set of 11200 labeled samples was used to evaluate the overall performance of the classifier.

Table \ref{tab:classifiers-results} shows that the use of computed residuals (rotational speed and current) as features for classification lead to improved results compared to a full four-feature vector. This can be explained by additional information introduced by the nominal model, including a small measure of difference between what is expected and what is actually being measured, leading to a better classification performance.

\begin{table}[!ht]
\centering\scriptsize
\caption{\label{tab:classifiers-results}Configurations for the evaluated classifiers. Training set: 12000 samples, 4-fold time series split cross-validation scheme. Test set: 11200 samples. Notation: L for LSTM cell, P for perceptron. Activation: tanh, recurrent activation: hard sigmoid, optimizer: ADAM, output layer activation: softmax.}
\begin{tabular}{ccccc}
\multicolumn{4}{c}{\textbf{Classifiers Configurations - Diagnosis}}\\
\hline
\textbf{Method} & \textbf{Batch Size} & \textbf{Hidden Layers} & \textbf{Avg. Accuracy}\\ 
\hline
$MLP_{all}$     & 5  & 48P,48P       & 73\% \\
$MLP_{res}$     & 5  & 48P,48P       & 75\% \\
$MLP_{all}$     & 1  & 48P,48P       & 74\% \\
$MLP_{res}$     & 1  & 48P,48P       & 78\% \\ 
$LSTM_{all}$    & 1  & 8L,48P,48P  & 69\% \\
$LSTM_{res}$    & 1  & 16L,48P,48P & 72\% \\
\hline 
\end{tabular}
\end{table}

The confusion matrix for the best performing classifier (MLP with residuals) is shown in Table \ref{tab:confusion-matrix-mlpres}. Our results show that identifying nominal operation is confused with a broken propeller, but identifying a broken propeller by itself works quite well. Biofouling is also hard to identify, with typical confusion with nominal operation. While we only achieve 78 \% accuracy on our dataset, these results also show that the problem of classifying soft-faults is hard, and there is lots of room for improvement.

\begin{table}[!ht]
\centering\scriptsize
\caption{\label{tab:confusion-matrix-mlpres}Normalized confusion matrix for the MLP classifiers using residuals.}
\begin{tabular}{ccccccc}
\multicolumn{7}{c}{\textbf{Confusion Matrix - MLP with Residuals}}\\
\hline
& \multicolumn{6}{c}{\textbf{Predicted labels}}\\
\cline{2-6}
\hline
Nominal     & 23 \% & 0 \% & 0 \% & 77 \% & 0 \% & 0 \%\\
13.0 V      & 0 \% & 100  \% & 0  \% & 0 \% & 0 \% & 0 \%\\
11.8 V      & 0 \% & 0 \% & 100 \% & 0 \% & 0 \% & 0 \%\\
Br. Prop.   & 6 \% & 0 \% & 0 \% & 94 \% & 0 \% & 0 \%\\
2 Br. Prop. & 4 \% & 0 \% & 0 \% & 82 \% & 14 \% & 0 \%\\
Biofouling  & 31 \% & 0 \% & 0 \% & 5 \% & 0 \% & 64 \%\\
\hline
\end{tabular}
\end{table}

\section{Conclusions and Future Work}
\label{sec:conlusions}

In this paper, we evaluate RNNs compared to the classic MLP for modeling and soft-fault classification of underwater thrusters using empirical data. As estimator, NARX network demonstrated to be able to identify more accurately the time dependencies presented by the thrusters. For soft-fault classification, we also evaluated LSTMs and MLPs.
Noncritical thruster failure conditions produce signal features capable of being diagnosed by a classifier-based approach, if the faulty behavior is previously known. In our experiments, a maximum of 78\% average accuracy score was obtained using the residuals computed from a nominal model. This result shows that modeling soft-faults is hard, as there are large confusions with other classes, indicating the need for further work.

Future work will include evaluation of this approach with data from real missions, and we will consider unbalanced datasets that are closer to real operations. We also expect that time-series models based on other kinds of features can improve our classification results.

\bibliography{phdlib}             
                                                   
\end{document}